\documentclass[letterpaper]{article} 
\usepackage{aaai25}  
\usepackage{times}  
\usepackage{helvet}  
\usepackage{courier}  
\usepackage[hyphens]{url}  
\usepackage{graphicx} 
\usepackage{natbib}  
\usepackage{caption} 
\usepackage{algorithm}
\usepackage{algorithmic}
\usepackage{newfloat}
\usepackage{listings}
\usepackage{booktabs}
\usepackage{amsmath}
\usepackage{amssymb}
\usepackage{color}
\usepackage{lettrine}
\usepackage{dsfont}
\usepackage{graphicx}
\usepackage{makecell}

\usepackage{newfloat}
\usepackage{listings}

\usepackage{xcolor}

\newcommand{\answerTODO}[1][]{\textcolor{red}{\bf [TODO]}}
\newcommand{\justificationTODO}[1][]{\textcolor{red}{\bf [TODO]}}

\DeclareCaptionStyle{ruled}{labelfont=normalfont,labelsep=colon,strut=off} 
\floatstyle{ruled}
\newfloat{listing}{tb}{lst}{}
\floatname{listing}{Listing}
\newcommand{\ptitle}[1]{\noindent\textbf{#1}\hspace{5pt}}

\title{Cross-PCR: A Robust Cross-Source Point Cloud Registration Framework}
\author{
    Guiyu Zhao, Zhentao Guo, Zewen Du, Hongbin Ma\thanks{Corresponding author} 
}
\affiliations{
    School of Automation, Beijing Institute of Technology\\

    zhaoguiyu@bit.edu.cn, \{zt\_guo1230, dzw1114\}@163.com, mathmhb@bit.edu.cn
}

\usepackage{bibentry}

\begin{document}

\maketitle

\begin{figure*}[htbp]
    \centering{\includegraphics[width=1.0\textwidth]{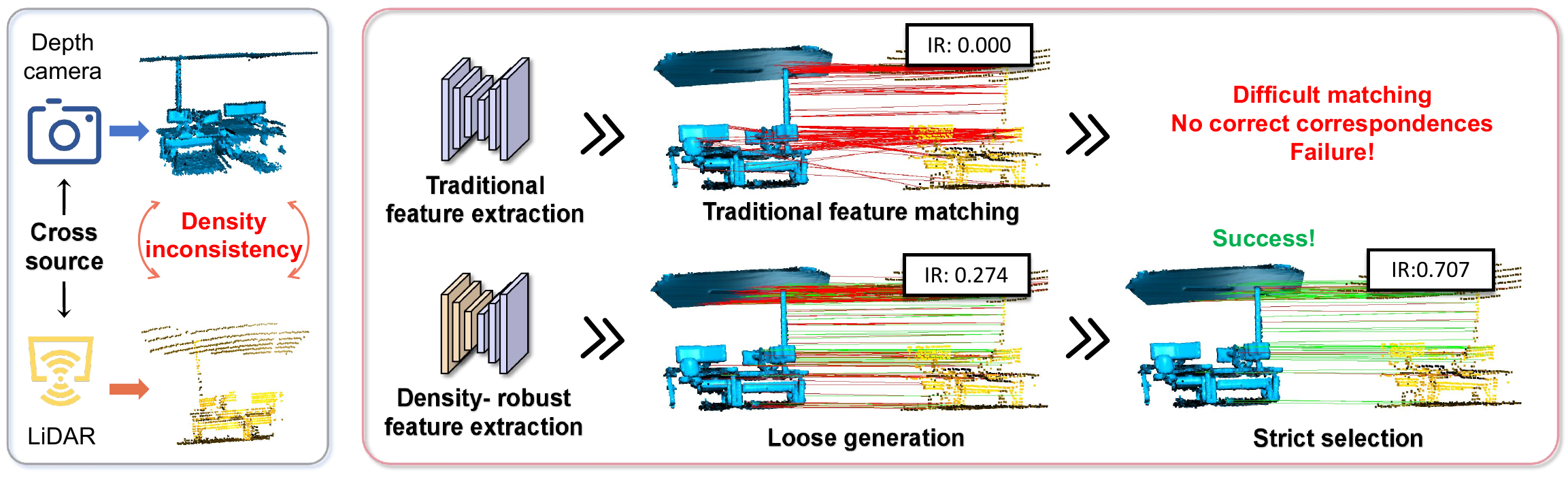}}  
    \vspace{-7pt}
    \caption{
    Existing methods often struggle with density differences and difficult matching, 
    leading to limited correct correspondences.
    In contrast, our Cross-PCR introduces a density-robust feature extractor to extract density-robust features. 
    Then, a loose-to-strict matching strategy, guided by the ``loose generation, strict selection'' idea, 
    generates reliable correspondences. 
        }
    \label{Introduction}
\end{figure*}

\begin{abstract}

Due to the density inconsistency and distribution difference 
between cross-source point clouds, previous methods fail in cross-source
point cloud
registration. We propose a density-robust feature 
extraction and matching scheme to achieve 
robust and accurate cross-source registration. 
To address the density inconsistency between cross-source data, 
we introduce a density-robust encoder for extracting density-robust features.
To tackle the issue of challenging feature matching and few correct correspondences, 
we adopt a loose-to-strict matching pipeline 
with a ``loose generation, strict selection'' idea. 
Under it,
we employ a one-to-many strategy to loosely generate initial correspondences. 
Subsequently, high-quality correspondences are strictly selected to 
achieve robust registration through sparse matching and dense matching. 
On the challenging Kinect-LiDAR scene in the cross-source 3DCSR dataset, 
our method improves feature matching recall by 63.5 percentage points (pp) 
and registration recall by 57.6 pp. 
It also achieves the best performance on 3DMatch, 
while maintaining robustness under diverse downsampling densities.
    
\end{abstract}

\section{Introduction}
\label{sec:intro}

Cross-source point cloud registration~\cite{huang2023cross, huang2021comprehensive} entails the 
alignment of point clouds obtained from diverse 
sensors into a shared coordinate system. 
It is the key to complete the calibration of different sensors for multi-sensor fusion.
Currently,
a variety of sensors and methods are employed 
for capturing point clouds, each offering distinct 
advantages in the data collection process. 
Consequently, harnessing the 
complementary strengths of different sensors for 
multi-sensor fusion holds great promise and finds 
applications in domains such as mapping, 
robotics, and remote sensing. 
In robotics, it is a cheap and feasible solution for localization.
We build a map with 
high-precision LiDAR and then scan the environment with inexpensive
sensors on the robot~\cite{xiong2023speal, CorrI2P}. 
Regrettably, cross-source
registration has seen limited advancements in recent years.

    Cross-source point cloud registration is 
    a notably challenging task, as it entails 
    aligning point clouds originating from distinct 
    sensors. These 
    discrepancies in point cloud collection 
    methodologies result in significant variations 
    in point cloud distribution, thereby introducing 
    formidable obstacles to achieving robust 
    registration. Therefore, cross-source point 
    cloud registration is beset by two primary challenges~\cite{huang2023cross}.
    \textbf{First}, as shown in Figure~\ref{Introduction}, there exist substantial disparities in the 
    distribution and density between two point clouds 
    from different sensors. For instance, LiDAR-generated point 
    clouds manifest as sparse line-like structures, while 
    those collected by depth cameras exhibit high density. 
    This divergence poses a large hindrance to the 
    extraction of similar features at corresponding points, 
    leading to a failure in correct matching (Figure~\ref{Introduction} right).
    Although DIF-PCR~\cite{liu2023density} proposes the density-robust feature, 
    it primarily addresses the density inequality within one point cloud, 
    rather than the density difference between two cross-source point clouds.
    \textbf{Second}, Different sensing 
    mechanisms~\cite{mallick2014characterizations, roriz2021automotive} introduces lots of different noises and outliers 
    into the cross-source point cloud data.
    These disturbances further exacerbate the difficulty of matching features.
    Consequently, only a limited number 
    of accurate correspondences can be obtained, 
    which also impedes the attainment of robust registration.
    
    Currently, the majority of point cloud registration 
    methods are primarily designed for homologous 
    data.  These methods encounter 
    challenges in addressing two 
    issues. 
    Unfortunately,
    although some methods~\cite{ao2021spinnet,zhao2023spherenet} 
    have good generalization 
    ability, they perform badly in cross-source registration tasks 
    with different densities between two point clouds. 
    SPEAL~\cite{xiong2023speal} proposes an outdoor cross-source 
    registration method, but it faces challenges in addressing 
    difficult matching issues in indoor scenes and fails to achieve robust registration.
    Due to the above two problems, 
    it is difficult to achieve cross-source registration by previous feature-based methods.
    Consequently, a considerable proportion of cross-source 
    point cloud registration algorithms rely on 
    optimization-based methodologies~\cite{zhao2023accurate, huang2023cross}. However, 
    these techniques face hurdles in achieving 
    robust registration across diverse scenarios.
    

    We introduce a feature-based cross-source
    registration method named Cross-PCR. To deal with the challenges 
    above, we present a density-robust feature extraction method
    and a loose-to-strict matching pipeline 
    with the idea of ``loose generation, strict selection''. 
    \textbf{For the first challenge}, 
    We propose a \textbf{density-robust feature extraction} network to address 
    density inconsistencies between cross-source point clouds.  
    Specifically, we design a multi-density fusion block to fuse the upsampled, 
    unsampled, and downsampled point cloud features respectively.  It is then 
    integrated into each layer of the encoder to achieve deep fusion of density,
    capturing common structural features between point clouds of different densities.  
    This greatly reduces the feature differences of cross-source point clouds in density.
    In addition, this simple and efficient design avoids large computational 
    overhead and complex training strategy~\cite{liu2023density}.
    \textbf{For the second one}, we devise a \textbf{loose-to-strict matching} 
    framework involving sparse matching and dense matching.
    Despite extracting the density-robust feature, 
    obtaining the most similar features at the corresponding points 
    remains challenging 
    in the context of cross-source registration. 
    Our key idea is ``loose generation, strict selection''
    which means loosely generating a large number of correspondences 
    and strictly filtering out correct correspondences. 
    To realize this, we propose a one-to-many matching strategy, 
    which aims to capture as many correct correspondences as possible. 
    Subsequently, the correspondences are systematically 
    pruned, facilitating the completion of sparse matching 
    through a robust consistency-based correspondence 
    refinement mechanism.
    Despite these efforts, results after sparse matching 
    may still contain spurious correspondences. 
    We propose a 
    prior-guided dense matching approach, 
    which globally filters dense correspondences in the feature space.
    It finally results in achieving a 
    more robust registration.

    We evaluate our Cross-PCR on both cross-source
    and same-source datasets. 
    On the 3DCSR dataset~\cite{huang2021comprehensive}, 
    we achieve the best performance with great improvement. 
    Our 
    method successfully achieves the 
    challenging registration 
    from depth camera to LiDAR,
    with a 57.6 percentage point (pp) improvement 
    in registration recall (RR) and 63.5 pp in feature matching recall (FMR). 
    Notably, we also achieve the 
    best results on 3DMatch~\cite{zeng20173dmatch}, 
    obtaining a remarkable 94.5\% RR and 88.7\% IR.
    Our main contributions are summarized as:
\begin{itemize}
    \setlength{\itemsep}{0pt}
    \setlength{\parsep}{0pt}
    \setlength{\parskip}{0pt}
    \item[$\bullet$] We introduce a novel
    loose-to-strict matching pipeline
    with the idea of ``loose generation, strict selection'',
    solving the problem of difficult feature matching in cross-source registration, 
    even with an extremely low IR.
    \item[$\bullet$] 
    We propose a density-robust feature extractor to extract the density-robust features, 
    addressing differences in density across cross-source point clouds.

    \item[$\bullet$]  
    Our method not only achieves robust cross-source registration,
    but it also exhibits effectiveness in same-source registration.  Moreover,
    it is the first to achieve robust registration on the
    indoor Kinect-LiDAR benchmark.

\end{itemize}


\section{Related Work}

\ptitle{Same-Source Registration.}
Recently, same-source point cloud registration has advanced rapidly. Early methods primarily relied on ICP optimization~\cite{segal2009generalized, sharp2002icp, yang2015go, yang2013go} and feature matching with handcrafted features~\cite{FPFH, PFH, salti2014shot}. Recent advancements, however, have shown that learning-based methods~\cite{zeng20173dmatch, ao2021spinnet, choy2019fully, huang2021predator, yu2021cofinet, qin2022geometric} offer more accurate and robust registration.
Among these, patch-based methods~\cite{poiesi2022learning, ao2021spinnet, ao2023buffer} demonstrate strong generalization, while fragment-based methods~\cite{choy2019fully, huang2021predator, yu2021cofinet} provide better efficiency. A key trend is the adoption of coarse-to-fine strategies~\cite{yu2021cofinet, yang2022one, lin2023coarse}, improving registration accuracy. Transformer-based methods~\cite{qin2022geometric, jin2024multiway} have become state-of-the-art, 
leveraging geometric information~\cite{qin2022geometric, yu2023rotation} and positional encoding~\cite{qin2022geometric, yang2022one, li2022lepard} to achieve outstanding results.

\ptitle{Cross-Source Registration.}
In contrast to same-source point cloud registration, the development of cross-source registration~\cite{huang2023cross, zhao2023accurate, huang2017coarse} has been relatively slow, primarily due to variations in point cloud distribution and density. The challenge of extracting similar features at corresponding points has limited feature-based methods, leading to a reliance on optimization~\cite{yang2013go, tazir2018cicp, zhao2023accurate} and model-based methods~\cite{huang2017coarse, huang2019fast, ling2022graph}.
Huang et al.~\cite{huang2017coarse} employ a coarse-to-fine strategy using ESF descriptors for 
cross-source registration. Zhao et al.~\cite{zhao2023accurate} achieve superior results through fuzzy 
clustering~\cite{gath1989unsupervised} and ICP~\cite{segal2009generalized}. SPEAL~\cite{xiong2023speal} 
uses skeleton point features for outdoor cross-source registration but performs poorly with LiDAR and 
depth camera data. 
Due to the lack of large cross-source indoor datasets, 
there are currently no effective learning-based methods~\cite{huang2020feature} for achieving robust indoor cross-source registration.


\begin{figure*}[t]
    \centering{\includegraphics[width=0.98\textwidth]{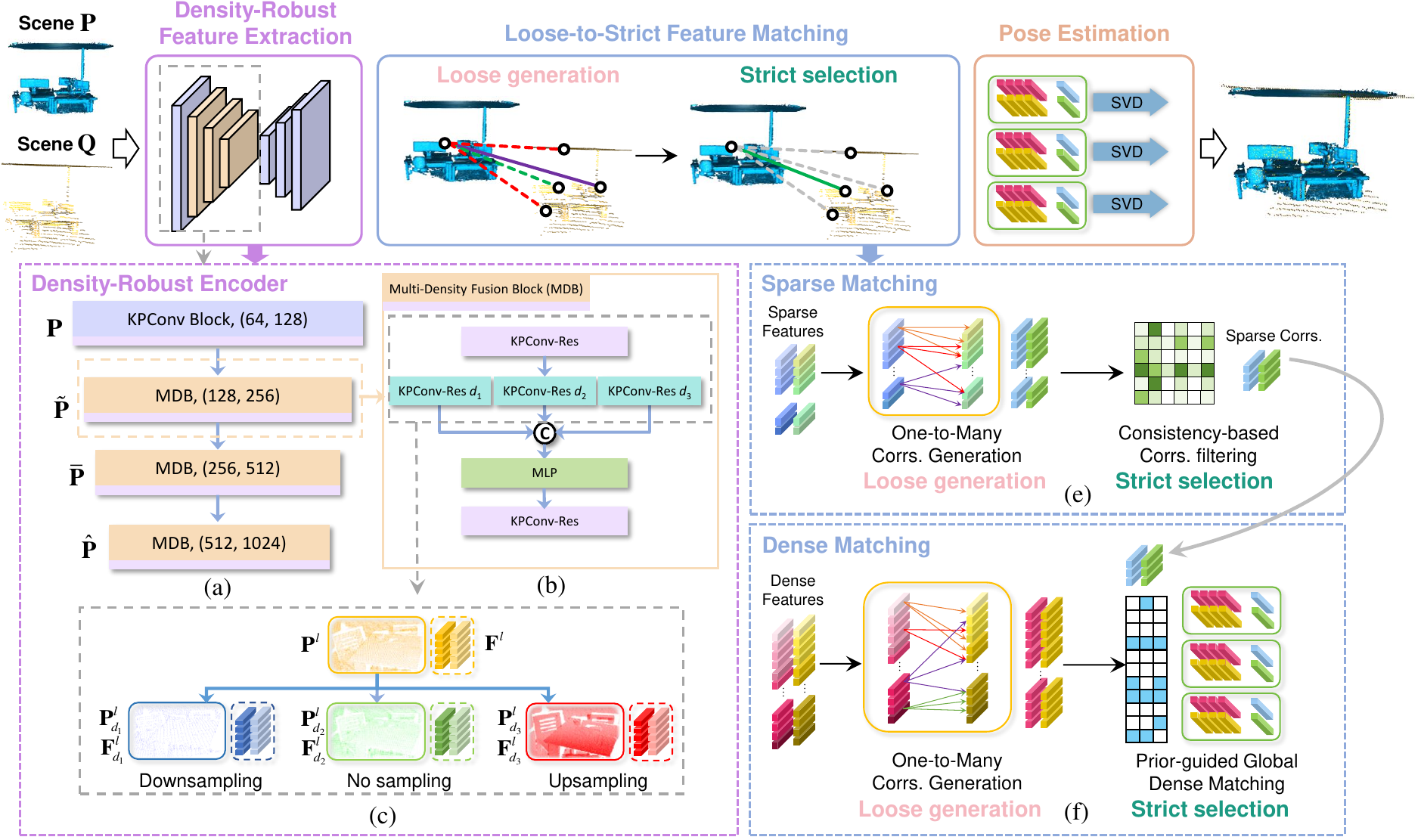}}  
    \caption{
        Our Cross-PCR consists of three parts: density-robust feature extraction, loose-to-strict feature matching, 
        and pose estimation. First, in density-robust feature extraction, 
        a pair of cross-source point clouds $\mathbf{P}$ and $\mathbf{Q}$ are fed into a 
        density-robust encoder to obtain the density-robust features $\mathbf{\widetilde{F} }^{\mathcal{{P}}}$ and
        $\mathbf{\widetilde{F} }^{\mathcal{{Q}}}$. After that, 
        in loose-to-strict feature matching, we propose a two-stage matching, which follows a 
        ``loose generation, strict selection'' strategy to generate $m$ sets of robust and 
        reliable correspondences $\mathcal{\widetilde {G}}_i^{\prime}$. 
        Finally, in pose estimation, we propose a novel hypothesis selection 
        method to select the best transformation $\mathbf{{T}}^{*}$.
    }
    \label{fig1}
    \vspace{-10pt}
\end{figure*}

\section{Method}
\label{sec:Method}

\ptitle{Problem Statement}
Cross-source point cloud registration aims to 
determine a rigid transformation $\mathbf T = \{\mathbf R,\mathbf t\}$, 
where $\mathbf R \in SO(3)$ represents a rotation matrix, and $\mathbf t \in \mathbb{R}^3$ 
represents a translation vector. 
The task is formulated as:
\begin{equation}
    \underset{\mathbf{R}\in SO(3), \mathbf{t}\in \mathbb{R}^3}{\arg \min } \sum_{\left({\mathbf{p}}_{x_i}, {\mathbf{q}}_{y_i}\right) \in \mathcal{G}} \left\|\mathbf{R} \cdot {\mathbf{p}}_{x_i}+\mathbf{t}-{\mathbf{q}}_{y_i}\right\|_2^2,
\end{equation}
where $\mathcal{G}$ represents the set of correspondences,
and $g_i = ({\mathbf{p}}_{x_i}, {\mathbf{q}}_{y_i})$ denotes a pair of corresponding points. 
As the correspondences are initially unknown, it is imperative to 
establish these correspondences through feature extraction and matching 
before solving the optimization problem. 

\subsection{Density-Robust Feature Extraction}\label{Extraction}


\ptitle{Density-Robust Encoder.}
Previously, the feature extraction network, employing KPConv-FPN~\cite{thomas2019kpconv, lin2017feature} 
as the backbone, leverages the advantageous characteristics of the 
FPN structure to ensure robust scale invariance in the features.   
However, in cross-source registration, the primary challenge 
lies in the density inconsistency between two point clouds.   
Unfortunately, the KPConv-FPN network 
is incapable of addressing this challenge, ultimately resulting 
in the failure to achieve robust cross-source registration.
Therefore,
we adopt a multi-density fusion block to devise a 
density-robust encoder (DRE), facilitating the capture of density-robust features.

Taking point cloud $\mathbf{P}$ as an example, we get different levels of point clouds $\{\mathbf{P}, \mathbf{\widetilde{P}}, \mathbf{\overline{P}},\mathbf{\widehat{P}}\}$ 
under the sampling strategy of KPConv~\cite{thomas2019kpconv}, where $\mathbf{\widetilde{P}}$ and $\mathbf{\widehat{P}}$ is the dense point cloud and the sparse point cloud, respectively. 
Then, 
as shown in Figure~\ref{fig1} (b), our density-robust encoder 
achieves hierarchical density-robust feature extraction 
on different levels.
It is constructed by one KPConv block 
and three multi-density fusion blocks (MDB).
For simplicity, we denote the $l$-level input feature 
and point cloud as $\mathbf{F}^{l}$ and $\mathbf{P}^{l}$.  
In a MDB, by employing the farthest point sampling (FPS)~\cite{qi2017pointnet++} and upsampling method~\cite{qi2017pointnet++} 
on $\mathbf{F}^{l}$ and $\mathbf{P}^{l}$, 
we generate three point clouds $\{\mathbf{P}_{d_1}^{l},\mathbf{P}^{l}_{d_2}, \mathbf{P}^{l}_{d_3}\}$ 
of varying densities along with their 
corresponding features $\{\mathbf{F}_{d_1}^{l},\mathbf{F}^{l}_{d_2}, \mathbf{F}^{l}_{d_3}\}$ (see Figure~\ref{fig1} (c)),
\begin{equation}
    \begin{aligned}
        \mathbf{F}_{d_1}^{l} &= \varPsi(\text{FPS}(\varPsi(\mathbf{F}^{l};\mathbf{P}^{l}))), \\
        \mathbf{F}_{d_2}^{l} &= \varPsi^2(\mathbf{F}^{l};\mathbf{P}^{l}), \\
        \mathbf{F}_{d_3}^{l} &= \varPsi(\text{UP}(\varPsi(\mathbf{F}^{l};\mathbf{P}^{l}))),
    \end{aligned}
\end{equation}
where $\varPsi(\mathbf{F}^{l};\mathbf{P}^{l})$ represents 
performing KPConv operator on $\mathbf{F}^{l}$
within $\mathbf{P}^{l}$.
$\text{FPS}(\cdot)$ and $\text{UP}(\cdot)$ denote downsampling and upsampling the point features, respectively.
The feature fusion of different densities is accomplished through multi-layer perceptron (MLP), and
the intermediate feature $\mathbf{F}^{l}_{m}$ after fusion is obtained:
\begin{equation}
    \mathbf{F}^{l}_{m} =\text{MLP}(\text{Cat}[
        \mathbf{F}_{d_1}^{l}, \mathbf{F}_{d_2}^{l}, \mathbf{F}_{d_3}^{l} ]),
\end{equation}
where concatenation operation and MLP are denoted as 
$\text{Cat}$ and $\text{MLP}(\cdot)$, respectively.
Finally, the input feature $\mathbf{F}^{l+1}$ of the next block is calculated
as $\mathbf{F}^{l+1} =\varPsi (\mathbf{F}^{l}_{m}; \mathbf{P}^{l+1})$.
By our density-robust encoder, we extract the 
density-robust features  $\mathbf{F}^{\mathcal{\widehat{P}}}, \mathbf{F}^{\mathcal{\widehat{Q}}}  \in \mathbb{R}^{|\mathcal{\widehat{P}}| \times {d}}$ of 
sparse points $\mathbf{\widehat{P}}$ 
and $\mathbf{\widehat{Q}}$.
And features $\mathbf{F}^{\mathcal{\widetilde{P}}}$ and $\mathbf{F}^{\mathcal{\widetilde{Q}}}$ 
of the dense points $\mathbf{\widetilde{P}}$ and $\mathbf{\widetilde{Q}}$ is obtained by KPConv decoder~\cite{bai2020d3feat}.

\ptitle{Attention Backbone.}
We refine the sparse features $\mathbf{F}^{\mathcal{\widehat{P}}}$ 
and $\mathbf{F}^{\mathcal{\widehat{Q}}}$ by employing the attention backbone~\cite{vaswani2017attention,qin2022geometric}. 
Our approach effectively 
captures long-distance dependencies
through the self-attention and cross-attention mechanisms. 
The refined sparse features are denoted as $\mathbf{\widetilde{F} }^{\mathcal{\widehat{P}}} \in \mathbb{R}^{|\mathcal{\widehat{P}}| \times \widetilde{d}}$ 
and $\mathbf{\widetilde{F} }^{\mathcal{\widehat{Q}}} \in \mathbb{R}^{|\mathcal{\widehat{Q}}| \times \widetilde{d}} $.

\subsection{Sparse Matching}

Despite our efforts to extract density-robust features, acquiring easily-matched features at corresponding points remains challenging. Figure~\ref{Introduction} shows the differences between point clouds from depth cameras and LiDAR. These disparities hinder the extraction of features that satisfy the nearest-neighbor relationship at corresponding points. Consequently, traditional feature matching in cross-source registration often results in few or no correct correspondences, leading to registration failures.
To address this, we introduce the ``loose generation, strict selection'' strategy. We propose a one-to-many correspondences generation method to produce broader and looser correspondences, maximizing the potential correct correspondences. Then, our correspondences filtering module 
selects the correct correspondences.

\ptitle{One-to-Many Correspondences Generation.}
First, we normalize the sparse features $\mathbf{\widetilde{F} }^{\mathcal{\widehat{P}}}$ 
and $\mathbf{\widetilde{F} }^{\mathcal{\widehat{Q}}}$, 
followed by calculating the feature similarity matrix $\mathbf{D}^{\mathbf{F}}=\text{Norm}(\mathbf{\widetilde{F} }^{\mathcal{\widehat{P}}}) \cdot \text{Norm}(\mathbf{\widetilde{F} }^{\mathcal{\widehat{Q}}})^\top $
where $\text{Norm}(\cdot)$ is a L2-normalization operation.
Utilizing the one-to-many strategy, 
for each source sparse point $\mathbf{\widehat{p}}_i$, we search the 
$k$ nearest neighbors within point cloud $\mathbf{\widehat{Q}}$ in the feature space,
through Top-k selection:


\begin{equation}
    \mathcal{\widehat{G}}_{i}=\Big\{ (\mathbf{\widehat{p}}_i,\mathbf{\widehat{q}}_{m_i}) | {m_i} \in \underset{j\in [1,|\mathbf{\widehat{Q}}|]}{{\text{Topk}} } (-\mathbf{D}_{i,j}^{\mathbf{F}}) \Big\},
\end{equation}
where ${\text{Topk}}(\cdot)$ is the operation 
that finds the index of $k$ maximum values. 
Finally, the correspondences $\mathcal{\widehat{G}}_{i}$ of each point  $\mathbf{\widehat{p}}_i$ 
is combined into the initial sparse correspondences 
$\mathcal{\widehat{G}}=\bigcup_{i=1}^{n} \mathcal{\widehat{G}}_{i}$
where $n$ is the number of points in $\mathbf{\widehat{P}}$.

\ptitle{Consistency-based Correspondences Filtering.}
Though our one-to-many correspondences generation strategy effectively 
captures a greater number of correct correspondences, 
it also introduces a substantial number of outliers 
due to its looser conditions, resulting in a 
lower inlier ratio. However, in cases where 
the inlier ratio is extremely low, the estimators~\cite{fischler1981random, ICP} 
become ineffective. 
Consequently, 
a strong correspondences filtering module is essential.

To improve efficiency, we use the spectral
matching technique~\cite{leordeanu2005spectral} to screen key sparse 
correspondences $\mathcal{\widehat{G}}^{k}$ on $\mathcal{\widehat{G}}$ where $\mathcal{\widehat{G}}^{k} \subseteq \mathcal{\widehat{G}}$.
We compute translation and rotation invariant measurements (TRIMs)~\cite{yang2020teaser} 
to assess the consistency distance $d_{ij}$ between the two 
correspondences ${\widehat{g}}_{i}^{k} \in \mathcal{\widehat{G}}^{k}$ and ${\widehat{g}}_{j} \in \mathcal{\widehat{G}}$
\begin{equation}
    d_{ij}= \Big| ||\mathbf{\widehat{p}}_i-\mathbf{\widehat{p}}_j||_2-||\mathbf{\widehat{q}}_i-\mathbf{\widehat{q}}_j||_2 \Big|
\label{eq1}
\end{equation}
where ${\widehat{g}}_{i}^{k}= (\mathbf{\widehat{p}}_i, \mathbf{\widehat{q}}_i)$ 
and ${\widehat{g}}_{j}= (\mathbf{\widehat{p}}_j, \mathbf{\widehat{q}}_j)$.
A smaller distance $d_{ij}$ indicates that the correspondence ${\widehat{g}}_{i}^{k}$ and ${\widehat{g}}_{j}$ satisfy a stronger consistency.
Given a threshold $\sigma_d$,
we get the consistency score between ${\widehat{g}}_{i}^{k}$ and ${\widehat{g}}_{j}$, 
denoted as $s_{ij} = \mathds {1}(d_{ij}  \leqslant {\sigma_d})$
where $\mathds {1}(\cdot)$ is the indicator function
and consistency score matrix is denoted as 
$\mathbf{S}_k=[s_{ij}] \in \mathbb{R}^{|\mathcal{\widehat{G}}^{k}| \times |\mathcal{\widehat{G}}|}$.
In the same way, we obtain consistency 
score matrix $\mathbf{S}_k \! \in\! \mathbb{R}^{|\mathcal{\widehat{G}}^{k}| \times |\mathcal{\widehat{G}}^{k}|}$ 
between key sparse correspondences $\mathcal{\widehat{G}}^{k}$.

To enhance the robustness of correspondence selection, 
inspired by $\text{SC}^2$ measure~\cite{chen2022sc2}, we 
calculate the weighted second-order consistency matrix  
$\mathbf{S}^{*}_s = \mathbf{S}_s\odot (\mathbf{S}_k\mathbf{W}_k\mathbf{S}_s)$
where operator 
$\odot$ represents the element-wise product. 
The weight $w_{ij}$ in matrix $\mathbf{W}_k \in \mathbb{R}^{|\mathcal{\widehat{G}}^{k}| \times |\mathcal{\widehat{G}}^{k}|}$, 
is defined as $w_{ij}=\exp(-\frac{d_{ij}^2}{2\sigma_d^2})$.
The element $s^{*}_{ij} $ of matrix $\mathbf{S}^{*}_s$ represents 
the number of correspondences $\mathcal{\widehat{G}}^k$ 
that simultaneously satisfy consistency with correspondence 
${\widehat{g}}_{i}^{k}$ and $\widehat{g}_j$. 
Then, we employ Top-k selection to choose $k$ sparse correspondences 
$\mathcal{\widehat{G}}^{\prime}_i$ with the 
highest consistency scores:
\begin{equation}
    \mathcal{\widehat{G}}_{i}^{\prime}\!=\!\Big\{ \left( \mathbf{\widehat{p}}_{l_i}, \mathbf{\widehat{q}}_{l_i}\right) \Big| 
    l_i\!\in\! \underset{j\in [1,|\mathcal{\widehat{G}}|]}{{\text{Topk}} }( (\mathbf{S}^{*}_s)_{ij}), 
    \left( \mathbf{\widehat{p}}_{l_i}, \mathbf{\widehat{q}}_{l_i}\right) \!\in\! \mathcal{\widehat{G}}
    \Big\}.
\end{equation}
By merging each set of filtered correspondences, we get the refined sparse correspondences
$\mathcal{\widehat{G}}^{\prime}=\bigcup_{i=1}^{|\mathcal{\widehat{G}}^k|} \mathcal{\widehat{G}}_{i}^{\prime}$.



\begin{figure*}[t]
    \centering{\includegraphics[width=1.0\textwidth]{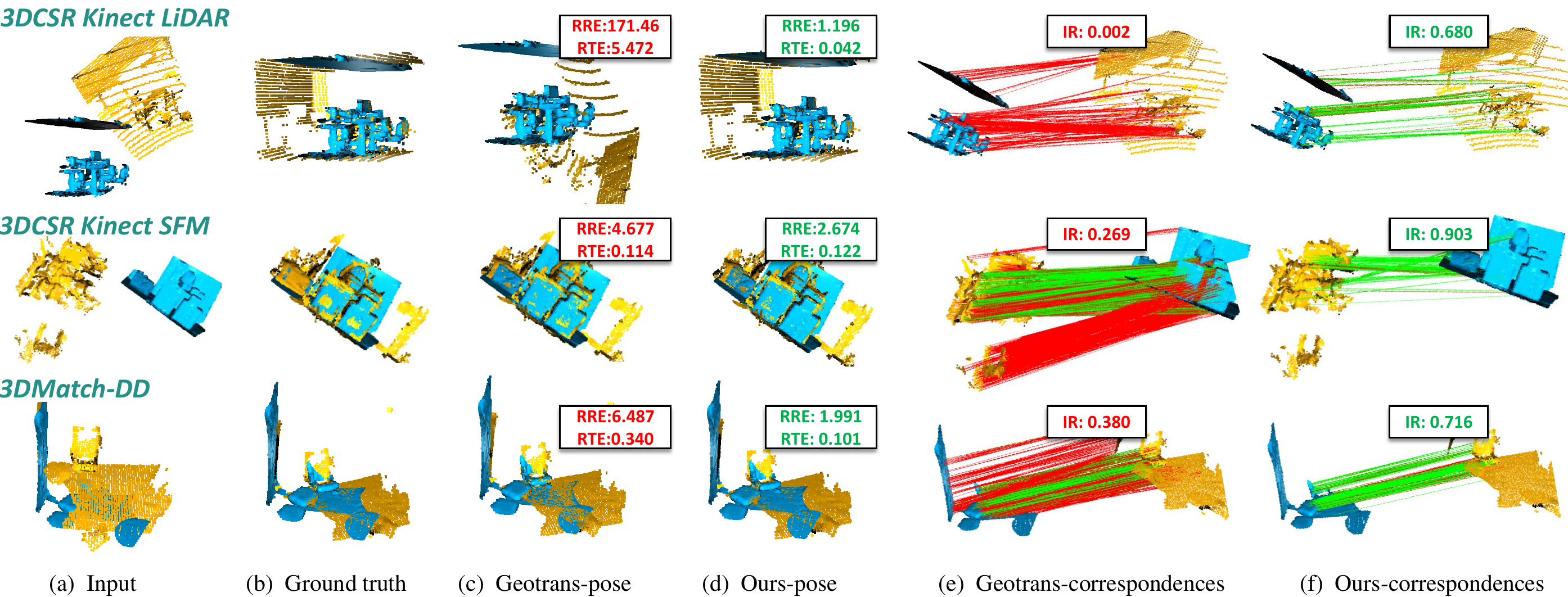}}  
    \vspace{-10pt}
    \caption{
        Correspondences and registration results compared with GeoTrans~\cite{qin2022geometric}}.
    \label{qr}
    \vspace{-5pt}
\end{figure*}

\subsection{Dense Matching}
Despite the refined sparse correspondences obtained through sparse matching, their limited number is insufficient for robust registration. However, their high quality allows us to use them as priors for guiding dense correspondences. Thus, we introduce prior-guided global dense matching based on these sparse correspondences.
Notably, some incorrect correspondences may still exist in the refined sparse matches. If using coarse-to-fine methods~\cite{qin2022geometric, yu2021cofinet}, incorrect sparse matches can misguide local dense points via optimal transport~\cite{sarlin2020superglue, yu2021cofinet}, causing widespread errors in dense correspondences.
To address this, we propose a more robust global-aware dense matching technique that transcends local matching constraints. Unlike coarse-to-fine strategies based on spatial neighbors, we adopt a global partitioning approach based on consistency.

Our general idea is to conduct a global search 
for every sparse correspondence by leveraging spatial consistency and 
aggregating dense correspondences that meet the consistency criteria.  
Each sparse correspondence ${\widehat{g}}^{\prime}_i$ in $\mathcal{\widehat{G}}^{\prime}$ 
yields its set of dense correspondences $\mathcal{\widetilde {G}}^{\prime}_i$ by Top-k selection.  

\ptitle{Prior-guided Global Dense Matching.}
Similar to sparse matching, we perform one-to-many correspondence generation 
on dense point features $\mathbf{F}^{\mathcal{\widetilde{P}}}$ and $\mathbf{F}^{\mathcal{\widetilde{Q}}}$, 
to establish loose dense correspondences $\mathcal{\widetilde{G}}$.
In the same way as Eq.~\ref{eq1}, we calculate the consistency matrix 
$\mathbf{S}_s^{\prime} \in \mathbb{R}^{|\mathcal{\widehat{G}}^{\prime}| \times |\mathcal{\widehat{G}}^{\prime}|}$ 
of the refined sparse correspondences $\mathcal{\widehat{G}}^{\prime}$.
Then, the sparse-to-dense consistency matrix $\mathbf{S}_{s2d}
\in \mathbb{R}^{|\mathcal{\widehat{G}}^{\prime}| \times |\mathcal{\widetilde{G}}|}$ 
between sparse correspondences $\mathcal{\widehat{G}}^{\prime}$  
and dense correspondences $\mathcal{\widetilde {G}}$ 
is calculated. The consistency score $s_{ij}$ in matrix $\mathbf{S}_{s2d}$
is defined as:
\begin{equation}
    s_{ij}= \mathds {1} \left(\Big| ||\mathbf{\widehat{p}}^{\prime}_i-\mathbf{\widetilde{p}}_j||_2-||\mathbf{\widehat{q}}^{\prime}_i-\mathbf{{\widetilde{q}}}_j||_2 \Big| \leqslant {\sigma_d}\right),
\end{equation}
where $(\mathbf{\widehat{p}}^{\prime}_i, \mathbf{\widehat{q}}^{\prime}_i) \in \mathcal{\widehat{G}}^{\prime}$,
$(\mathbf{{\widetilde{p}}}_j, \mathbf{{\widetilde{q}}}_j) \in \mathcal{\widetilde {G}}$. 
Then, we  
calculate the sparse-to-dense second-order consistency matrix $\mathbf{S}^{*}_{s2d}$
\begin{equation}  
    \mathbf{S}^{*}_{s2d} = \mathbf{S}_{s2d} \odot (\mathbf{S}_s^{\prime} \mathbf{W}_{s}^{\prime}\mathbf{S}_{s2d} )
\end{equation}
where $\mathbf{W}_s^{\prime} \in \mathbb{R}^{|\mathcal{\widehat{G}}^{\prime}| \times |\mathcal{\widehat{G}}^{\prime}|}$ is the weight matrix 
between refined sparse correspondences $\mathcal{\widehat{G}}^{\prime}$, and
its calculation is identical to that of $\mathbf{W}_s$.
Top-k is used to select the top $k$ dense correspondences $\mathcal{\widetilde {G}}_i^{\prime} \subseteq  \mathcal{\widetilde {G}}$
for each sparse correspondence ${\widehat{g}}^{\prime}_i$:
\begin{equation}
    \mathcal{\widetilde {G}}_i^{\prime}\!=\!\Big\{\! \left( \mathbf{\widetilde{p}}_{l_i}\!,\! \mathbf{\widetilde{q}}_{l_i}\right) \!\Big| 
    l_i\!\in\! \underset{j\in [1,|\mathcal{{G}}|]}{{\text{Topk}} }( (\mathbf{S}^{*}_{s2d})_{ij}),
    \left( \mathbf{\widetilde{p}}_{l_i}, \mathbf{\widetilde{q}}_{l_i}\right) \!\in\!  \mathcal{\widetilde {G}}
    \!\Big\}
\end{equation}
This results in $m$ groups of dense 
correspondences, completing the dense matching.
Finally, $m$ transformations $\mathbf T_i =\{\mathbf R_i, \mathbf t_i \}$ are obtained by weighted SVD~\cite{ICP} 
where the weight is the consistency score within each group 
of dense correspondences $\mathcal{\widetilde {G}}_i$.
The details of weighted SVD are provided in the Appendix.

\ptitle{Hypothesis Selection.}
Due to the low inlier ratio of the initial correspondences 
generated from the cross-source point clouds, 
it may not be feasible to accurately determine the optimal transformation 
by utilizing initial correspondences for hypothesis selection. 
Consequently, different from the previous method~\cite{chen2022sc2,qin2022geometric}, 
we choose to directly compute the chamfer-like truncated distance 
between the pair of sparse point clouds $\mathbf{\widehat{P}}$ and $\mathbf{\widehat{Q}}$.
The selection of optimal transformation $\mathbf{{T} }^{*}$ is 
completed by solving the following optimization problem:
\begin{equation}
    { \mathbf{R}}^{*}, { \mathbf{t}}^{*}=\max _{\mathbf{R}_i, \mathbf{t}_i} 
    \sum_{\mathbf{\widehat{p}}_n \in \mathbf{\widehat{P}}} 
    \mathds {1}\big(
        \min _{\mathbf{\widehat{q}}_m \in \mathbf{\widehat{Q}}}
        \left\|  \mathbf{\widehat{p}}_n-\mathbf{\widehat{q}}_m  \right\|< \tau_0 
        \big),
\end{equation}
where $\tau_0 $ is a distance threshold (\emph{i.e.} 0.1m).

\ptitle{Loss Function} 
The loss function $\mathcal{L}=\mathcal{L}_{s}+\mathcal{L}_{d}$ is composed of 
sparse matching loss and dense matching loss. 
To improve robustness to low overlap, the sparse matching loss uses 
the overlap-aware circle loss~\cite{qin2022geometric}.
For efficiency, so we only use circle loss~\cite{sun2020circle} to supervise metric learning of dense point features.



\section{Experiments}

We evaluate our Cross-PCR on both cross-source
dataset 3DCSR~\cite{huang2021comprehensive} 
and same-source dataset 3DMatch~\cite{zeng20173dmatch}.  
We also perform a robustness test
to highlight the robustness to difference in density.  

Following~\cite{yu2021cofinet,qin2022geometric}, we use five metrics to 
evaluate our performance: 
(1) \emph{Inlier Ratio} (IR), represents 
the fraction of
correspondences whose residuals is less than a threshold ($0.1$m).
(2) \emph{Feature Matching Recall} (FMR), represents the 
proportion of point cloud pairs whose IR 
is greater than 5\% in the total point cloud pairs. 
(3) \emph{Registration Recall} (RR)\footnote{
    Following~\cite{huang2021predator}, 
    for same-source registration, RMSE$<$0.2m is considered as successful 
    registration; Following~\cite{zhao2023accurate}, 
    for cross-source registration, RE$<$15$^{\circ}$ and TE$<$30cm is 
    considered as successful registration.
},
represents 
the proportion of correctly registered point cloud pairs to the 
total point cloud pairs. 
(4) \emph{Rotation Error} (RE), 
the geodesic distance between the estimated rotation 
and the ground-truth rotation. 
(5) \emph{Translation Error} (TE),
the Euclidean distance between the estimated translation 
and the ground-truth translation.

\begin{table}[t]
    \centering
    \renewcommand\arraystretch{0.6}
    \resizebox{1.0\linewidth}{!}{
        \begin{tabular}{l|ccccc}
            \toprule
            Method & RR($\uparrow$) & IR($\uparrow$) & FMR($\uparrow$) & RE($\downarrow$) & TE($\downarrow$)     \\ 
            \midrule
            \multicolumn{6}{c}{\emph{Kinect\_sfm}} \\
            \midrule
            GICP &12.5 &- &- &\textbf{1.90} &\textbf{0.03} 	\\
            JRMPC &0 &3.3 &14.8 &- &- \\
            GCTR\footnotemark{}  &15.2 &10.9 &46.8 &2.99 &0.10 	\\
            GCC\footnotemark[\value{footnote}] &81.3 &- &- &2.09 &\underline{0.06} \\
            SpinNet &\underline{84.5} &\underline{54.8} &\underline{93.8} &2.64 &0.08 	\\
            Predator &71.8 &31.8 &81.3 &3.93 &0.11 	\\
            CoFiNet &37.5 &25.3 &90.6&4.19 &0.10 \\
            GeoTransformer  &84.4 &40.2 &90.6 &\underline{1.95} &\underline{0.06} 	\\
            Cross-PCR (\emph{ours}) &\textbf{93.8} &\textbf{89.4} &\textbf{96.9} &{2.20} &\underline{0.06} 	\\
            \midrule 
            \multicolumn{6}{c}{\emph{Kinect\_lidar}} \\
            \midrule
            GICP &0.6 &- &- &12.8 &0.24  	\\
            JRMPC	&0 &0.5 &3.1 &- &- \\
            GCTR\footnotemark[\value{footnote}]	&0 &0.1 &2.1 &- &-  \\
            GCC\footnotemark[\value{footnote}]  	&3.1 &- &- &\underline{3.64} &\underline{0.15} \\
            SpinNet  	&5.2 &1.2 &7.1 &6.93 &\underline{0.15} \\
            Predator 	&5.2 &0.5 &1.3 & 4.63 &0.21 \\
            CoFiNet	&5.2 &1.2 &7.1 &4.65 &\underline{0.15} \\
            GeoTransformer	&\underline{9.1} &\underline{3.4} &\underline{9.7} &4.60 &0.17 \\
            Cross-PCR (\emph{ours})  	&\textbf{66.7} &\textbf{35.5} &\textbf{73.2} &\textbf{2.83} &\textbf{0.14} \\
            \bottomrule
    \end{tabular}
    }
    \vspace{-7pt}
    \caption{
        Quantitative results on 3DCSR. 
    }
    \label{3dmatch2}
    \vspace{-10pt}
\end{table}
\footnotetext{
As the codes are not available, we reproduce results ourselves.
}

\subsection{Evaluation on Cross-Source Dataset}\label{Cross-Source}

\ptitle{Dataset.}
3DCSR dataset~\cite{huang2021comprehensive} is an indoor cross-source 
dataset with two cross-source categories, 
containing 202 cross-source scenarios. The first  
category is \emph{Kinect-sfm}, which consists of the point clouds collected by the
\emph{Kinect camera} and the point clouds generated by the \emph{structure 
from motion} (SFM).
The second one is \emph{Kinect-LiDAR}, which consists of the 
point clouds collected by \emph{LiDAR device} and the 
point clouds collected by \emph{Kinect}.

\ptitle{\emph{Kinect-SFM} results.}
We conduct a comprehensive evaluation between our 
Cross-PCR and cross-source 
registration algorithms:
GICP~\cite{segal2009generalized},
JRMPC~\cite{huang2017coarse}, 
GCTR~\cite{huang2019fast},
GCC~\cite{zhao2023accurate}
and learning-based same-source 
registration methods: SpinNet~\cite{ao2021spinnet}, 
Predator~\cite{huang2021predator},  
CoFiNet~\cite{yu2021cofinet}, 
GeoTransformer \cite{qin2022geometric}.
The results on \emph{Kinect-SFM} are shown 
in Table~\ref{3dmatch2}.
Following~\cite{huang2019fast, zhao2023accurate}, 
we evaluate the registration results through RR, TE, and RE.
Additionally, to underscore 
the superior feature matching capability of our method, 
we evaluate the correspondences results 
using IR and FMR.
On the \emph{Kinect-SFM} benchmark, 
our approach performs best on RR, IR, and FMR. 
Notably, benefiting from our loose-to-strict matching,
our method demonstrates remarkable efficacy in feature matching, 
achieving a 34.6 pp improvement on IR, surpassing other 
algorithms by a considerable 
margin. 
Qualitative results are shown in Figure~\ref{qr}. 

\ptitle{\emph{Kinect-LiDAR} results.}
We also assess the outcomes on \emph{Kinect-LiDAR} in the same way, 
with the results presented in the right part of Table~\ref{3dmatch2}.  
Notably, cross-source data in the \emph{Kinect-LiDAR} benchmark 
exhibits greater dissimilarity than that in \emph{Kinect-SFM}.  
As indicated in Table~\ref{3dmatch2}, 
almost all methods fail to achieve correct registration on \emph{Kinect-LiDAR} benchmark. 
Benefiting from the idea of ``loose generation, strict selection'' and density-robust features, 
our method is the first to achieve robust registration 
on the \emph{Kinect-LiDAR} benchmark, with the lowest RE and TE.  
it achieves a 66.7\% RR, 35.5\% IR, and 73.2\% FMR, 
marking respective improvements of 57.6 pp, 32.1 pp, and 
63.5 pp compared to the previous SOTA method~\cite{qin2022geometric}.  
Our method exhibits a remarkable generalization ability across diverse modal data, 
ensuring robust cross-source registration.



\begin{table}[t]
    \setlength{\tabcolsep}{3pt}
    \renewcommand\arraystretch{0.8}
    \centering

    \resizebox{1.0\linewidth}{!}{
    \begin{tabular}{l|ccccc|ccccc}
    \toprule
        & \multicolumn{5}{c|}{3DMatch} & \multicolumn{5}{c}{3DLoMatch} \\
    \# Samples & 5000 & 2500 & 1000 & 500 & 250 & 5000 & 2500 & 1000 & 500 & 250 \\
    \midrule
    \multicolumn{11}{c}{\emph{Feature Matching Recall} (\%) $\uparrow$} \\
    \midrule
    SpinNet & 97.6 & 97.2 & 96.8 & 95.5 & 94.3 & 75.3 & 74.9 & 72.5 & 70.0 & 63.6 \\
    Predator& 96.6 & 96.6 & 96.5 & 96.3 & 96.5 & 78.6 & 77.4 & 76.3 & {75.7} & 75.3 \\
    YOHO & \textbf{98.2} & 97.6 & 97.5 & 97.7 & 96.0 & 79.4 & 78.1 & 76.3 & 73.8 & 69.1 \\
    CoFiNet & \underline{98.1} & \textbf{98.3} & \textbf{98.1} & \textbf{98.2} & \textbf{98.3} & {83.1} & {83.5} & {83.3} & {83.1} & {82.6} \\ 
    GeoTransformer& 97.9 & {97.9} & \underline{97.9} & {97.9} & {97.6} & \underline{88.3} & \underline{88.6} & \underline{88.8} & \underline{88.6} & \underline{88.3} \\ 
    RoITr&98.0  &\underline{98.0} &\underline{97.9}  &\underline{98.0} &\underline{97.9} &\textbf{89.6}  &\textbf{89.6} &\textbf{89.5}  &\textbf{89.4} &\textbf{89.3} \\
    Cross-PCR (\emph{ours})  &{97.9}  &{97.7} &\underline{97.9}  &{97.7} &{97.7} &83.5  &83.3 &83.3  &83.1 &83.4 \\

    \midrule
    \multicolumn{11}{c}{\emph{Inlier Ratio} (\%) $\uparrow$} \\
    \midrule
    SpinNet & 47.5 & 44.7 & 39.4 & 33.9 & 27.6 & 20.5 & 19.0 & 16.3 & 13.8 & 11.1 \\
    Predator& 58.0 & 58.4 & {57.1} & {54.1} & 49.3 & {26.7} & {28.1} & {28.3} & {27.5} & 25.8 \\
    YOHO & {64.4} & {60.7} & 55.7 & 46.4 & 41.2 & 25.9 & 23.3 & 22.6 & 18.2 & 15.0 \\
    CoFiNet  & 49.8 &  51.2 & {51.9} &  52.2 &  52.2 & {24.4} & {25.9} & {26.7} & {26.8} & {26.9} \\ 
    GeoTransformer  & {71.9} & {75.2} & {76.0} & {82.2} & \underline{85.1} & {43.5} & {45.3} & {46.2} & {52.9} & {57.7} \\
    RoITr &\underline{82.6}  &\underline{82.8} &\underline{83.0}  &\underline{83.0} &{83.0} &\underline{54.3}  &\underline{54.6} &\underline{55.1}  &\underline{55.2} &\underline{55.3} \\
    Cross-PCR (\emph{ours})   &\textbf{88.7}  &\textbf{88.7}  &\textbf{88.7}   &\textbf{88.7}  &\textbf{88.7}  &\textbf{65.9}  &\textbf{65.9} &\textbf{65.9}  &\textbf{65.9} &\textbf{65.9} \\
    \midrule
    \multicolumn{11}{c}{\emph{Registration Recall} (\%) $\uparrow$} \\
    \midrule
    SpinNet & 88.6 & 86.6 & 85.5 & 83.5 & 70.2 & 59.8 & 54.9 & 48.3 & 39.8 & 26.8 \\
    Predator & 89.0 & 89.9 & {90.6} & 88.5 & 86.6 & 59.8 & 61.2 & 62.4 & 60.8 & 58.1 \\
    YOHO & {90.8} & {90.3} & 89.1 & {88.6} & 84.5 & 65.2 & 65.5 & 63.2 & 56.5 & 48.0 \\
    CoFiNet  & 89.1 & 88.9 & {88.4} & {87.4} & {87.0} & {67.5} & {66.2} & {64.2} & {63.1} & {61.0} \\ 
    GeoTransformer & \underline{92.0} & \underline{91.8} & \underline{91.8} & \underline{91.4} & \underline{91.2} & \textbf{75.0} & \textbf{74.8} & \underline{74.2} & \underline{74.1} & {73.5} \\
    RoITr  &91.9  &91.7 &\underline{91.8}  &\underline{91.4} &91.0  &\underline{74.7}  &\textbf{74.8} &\textbf{74.8}  &\textbf{74.2} &\underline{73.6} \\
    Cross-PCR (\emph{ours}) &\textbf{94.5}  &\textbf{94.2} &\textbf{94.2}  &\textbf{94.3} &\textbf{94.0}  &73.7  &\underline{73.9} &74.1  &\textbf{74.2} &\textbf{74.1} \\
    \bottomrule
    \end{tabular}
    }
    \vspace{-7pt}
    \caption{
        Quantitative results on 3DMatch and 3DLoMatch.
        }
    \label{table:results-3dmatch}
    \vspace{-10pt}
    
\end{table}

\begin{table*}[ht]
    \renewcommand\arraystretch{0.8}
    \resizebox{1.0\linewidth}{!}{
    \centering
    \begin{tabular}{l|ccc|ccc|ccc|ccc|ccc|ccc}
    \toprule
    Benchmark& \multicolumn{9}{c|}{3DMatch} & \multicolumn{9}{c}{3DLoMatch} \\
    \cmidrule{1-1} \cmidrule(lr){2-19} 
    Voxel length & \multicolumn{3}{c}{-} & \multicolumn{3}{c}{0.05} & \multicolumn{3}{c|}{0.1} & \multicolumn{3}{c}{-} & \multicolumn{3}{c}{0.05} & \multicolumn{3}{c}{0.1} \\
    \cmidrule{1-1} \cmidrule(lr){2-4}  \cmidrule(lr){5-7} \cmidrule(lr){8-10} \cmidrule(lr){11-13} \cmidrule(lr){14-16} \cmidrule(lr){17-19}
    Metrics &RR &IR &FMR &RR &IR &FMR &RR &IR &FMR &RR &IR &FMR &RR &IR &FMR &RR &IR &FMR \\
    \midrule

    Predator&89.0&58.0&96.6  &2.1&0.4&1.1  &2.1&0.1&0.2
    &59.8&26.7&78.6  &0.2 &0&0  &\underline{2.1}&0.1&0.2 \\

    CoFiNet &89.1&49.8&\textbf{98.1} &2.0&0.4&1.5  &0.4&0.1&0  
    &67.5&24.4&83.1  &0.1&0& 0   &0.1&0&0  \\

    GeoTrans &\underline{92.0}&\underline{71.9}&\underline{97.9}  &\underline{91.1}&\underline{63.0}&\textbf{97.6} &\underline{4.3}&\underline{2.3}&\underline{14.3}
    &\textbf{75.0}&\underline{43.5}&\textbf{88.3}  &\underline{66.9}&\underline{33.5}&\textbf{83.3}   &1.7&\underline{1.2}&\underline{5.6} \\

    
    Cross-PCR (\emph{ours})&\textbf{94.5} &\textbf{88.7} &{97.7} &\textbf{91.5} &\textbf{84.4} &\underline{97.2} &\textbf{91.0} &\textbf{81.9} &\textbf{97.2}
    & \underline{73.7} &\textbf{65.9} &\underline{83.5} &\textbf{69.8}&\textbf{61.4}&\underline{80.2} &\textbf{65.8}&\textbf{57.2}&\textbf{77.4} \\
    \bottomrule
    \end{tabular}
    }
    \vspace{-7pt}
    \caption{
        Quantitative results on the 3DMatch-DD benchmark.
    }
    \label{table:arbitrary downsampling}
    \vspace{-4pt}
\end{table*}

\subsection{Evaluation on Same-Source Dataset}\label{3DMatch}

\ptitle{Dataset.}
The 3DMatch dataset~\cite{zeng20173dmatch} is a large indoor dataset containing 
62 scenarios, of which 46 are used for 
training, 8 for validation, and 8 for testing. 
Following~\cite{huang2021predator}, point cloud pairs 
with overlap $>30$\% are split as 3DMatch, 
and those with 10\% $\thicksim$ 30\%  overlap
are split as 3DLoMatch~\cite{huang2021predator}.

\ptitle{Correspondence results.}
Following~\cite{qin2022geometric, yu2023rotation}, we conduct 
experiments to assess the robustness of samples.
Comparative evaluations are made with the state-of-the-art methods: 
SpinNet~\cite{ao2021spinnet},
Predator~\cite{huang2021predator},
YOHO~\cite{wang2021you},
CoFiNet~\cite{yu2021cofinet}, 
GeoTrans~\cite{qin2022geometric}, RoITr~\cite{yu2023rotation}. Correspondence 
results are evaluated using FMR and IR, as shown 
in Table~\ref{table:results-3dmatch} (top and middle). 
For IR, our method exhibits significant enhancement, 
achieving 6.1 pp and 11.6 pp improvements over the state-of-the-art RoITr~\cite{yu2023rotation} 
on 3DMatch and 3DLoMatch, respectively. 
For FMR, our method also yields commendable results, 
but it is slightly worse than RoITr~\cite{yu2023rotation}.
The drop arises from the removal of ambiguous inliers 
through loose-to-strict matching, resulting in the correspondences 
with low IR ($<$5\%) in certain scenarios. 

\ptitle{Registration results.}
Following~\cite{qin2022geometric}, we employ the 
RR to assess our registration results. For other methods, we employ a 50k RANSAC 
to estimate transformation.  Compared to other methods, Cross-PCR yields 
the best outcomes in most cases, 
surpassing RoITr~\cite{yu2023rotation} by 2.6 pp.  
Our method also exhibits robustness to sampling, 
with only a 0.5 pp decrease in RR from 5000 to 250 samples.  
Moreover,   
instead of using time-expensive RANSAC, 
our method only uses weight SVD~\cite{ICP} which has a low time cost.
This underscores the high quality of the correspondences, 
enabling robust registration without relying on a 
robust estimator.


\begin{table}[t]
    \setlength{\tabcolsep}{3pt}
    \renewcommand\arraystretch{0.9}
    \centering
    \resizebox{1.0\linewidth}{!}{
    \begin{tabular}{l|ccc|ccc|ccc}
    \toprule

    Voxel length & \multicolumn{3}{c|}{-} & \multicolumn{3}{c|}{0.05} & \multicolumn{3}{c}{0.1} \\
    \midrule
    Metrics &RR($\uparrow$) &IR($\uparrow$) &FMR($\uparrow$) &RR($\uparrow$) &IR($\uparrow$) &FMR($\uparrow$) &RR($\uparrow$) &IR($\uparrow$) &FMR($\uparrow$)  \\
    \midrule
    KPConv &92.2&87.4&96.5  &90.7&\textbf{85.5} &96.2   &57.6&40.9&69.2 
    \\
    DRE*&\textbf{94.5} &\textbf{88.7} &\textbf{97.7} &\textbf{91.5} &{84.4} &\textbf{97.2} &\textbf{91.0} &\textbf{81.9} &\textbf{97.2}
    \\
    \bottomrule
    \end{tabular}
    }
    \vspace{-7pt}
    \caption{
        Ablation study with DRE on 3DMatch-DD.
    }
    \label{Ablation2}
    \vspace{-10pt}
\end{table}

\subsection{Robustness to Density Inconsistency}\label{Robustness}

To further evaluate the robustness to density inconsistency,
we downsample the 3DMatch 
dataset to various extents, creating 
the 3DMatch-DD benchmark.  

\ptitle{Experiments settings.}
Following~\cite{huang2021predator}, 
we preprocess the original 
3DMatch dataset. Then, 
we only downsample the target point cloud by
using \emph{voxel downsample}
to simulate the large 
density difference between the cross-source point cloud pair. 
By setting no voxel downsampling, voxel downsampling with a 0.05m side 
length, and with a 0.1m side length, 
we conduct three experiments. 

\ptitle{Quantitative results.}
Table~\ref{table:arbitrary downsampling} presents quantitative results. 
On the 3DMatch-DD benchmark, 
our method consistently outperforms across diverse downsampling levels
compared to 
other baselines~\cite{huang2021predator,yu2021cofinet,qin2022geometric}. 
In the challenging condition 
of voxel downsampling 
with a 0.1m side length,
our method still achieves correct registration in certain scenarios. 
It only drops 3.5 pp and 7.9 pp in RR compared to the original, 
far less than GeoTrans~\cite{qin2022geometric}, which drops 86.7 pp and 73.3 pp.

\begin{table}[t]
    \renewcommand\arraystretch{0.8}
    \centering
    \resizebox{1.0\linewidth}{!}{
    \begin{tabular}{l|l|ccc|ccc}
    \toprule
    & &\multicolumn{3}{c|}{\emph{Kinect-sfm}} & \multicolumn{3}{c}{\emph{Kinect-lidar}} \\
    No.  &Methods        & RR($\uparrow$)  & IR($\uparrow$) & FMR($\uparrow$) & RR($\uparrow$)  & IR($\uparrow$) & FMR($\uparrow$) \\
    \midrule
    
    1)         &KPConv        &87.5&85.2   &{93.8}   &57.7 &30.1  &61.7\\
    2)         &DRE*     &\textbf{93.8} &\textbf{89.4} &\textbf{96.9} &\textbf{66.7}  &\textbf{35.5} &\textbf{73.2}\\
    \midrule  
    3)         &W/o loose generation        &84.5   &84.4  &93.8 &44.2  &26.4  &54.5\\
    4)         &W/ loose generation*   &\textbf{93.8} &\textbf{89.4} &\textbf{96.9} &\textbf{66.7} &\textbf{35.5} &\textbf{73.2}\\
    \midrule  
    5)         &W/o strict selection       &81.1   &77.9   &90.6 &27.3  &17.5  &34.4\\
    6)         &W/ strict selection*        &\textbf{93.8} &\textbf{89.4} &\textbf{96.9} &\textbf{66.7} &\textbf{35.5} &\textbf{73.2}\\
    \midrule  
    7)         &One-stage  matching     &68.8   &29.5   &93.8 &56.2 &10.3 &50.9  \\
    8)         &Two-stage matching*       &\textbf{93.8} &\textbf{89.4} &\textbf{96.9} &\textbf{66.7}  &\textbf{35.5} &\textbf{73.2}\\
    \bottomrule
    \end{tabular}
    }
    \vspace{-7pt}
    \caption{Ablation study on the 3DCSR dataset. }
    \label{Ablation}
    \vspace{-10pt}
    \end{table}
\subsection{Ablation Study}
\label{AblationS}

\ptitle{Density-robust encoder.}
We substitute our density-robust encoder with KPConv-FPN~\cite{qin2022geometric}. The results are shown in Table~\ref{Ablation}, entries (1) and (2). For the \emph{Kinect-lidar} benchmark, the ablation model performance decreases significantly due to large density differences. Further ablation on the 3DMatch-DD benchmark (Table~\ref{Ablation2}) shows that our method has much less performance degradation across various sampling levels compared to the ablation model.

\ptitle{W/o loose generation.}
We replace our one-to-many correspondences generation with 
traditional one-to-one matching. As shown in 
Table~\ref{Ablation} (3) and (4), there is a substantial 
improvement on 
\emph{Kinect-lidar}, with ``loose generation''. 
This shows the technique contributes to achieving 
robust registration even when matching is extremely challenging.

\ptitle{W/o strict selection.}
As shown in Table~\ref{Ablation} (5) and (6), we remove the 
consistency-based correspondences refinement module. 
The ablated model achieves 
low RR and IR, 
which demonstrates this module is critical to achieving robust registration.
Entries (3) and (5) together prove the superiority of our ``loose generation, strict selection'' idea.


\ptitle{One-stage vs two-stage.}
We only reserve sparse matching for one-stage matching to compare with our two-stage matching. 
As shown in Table~\ref{Ablation} (7) and (8), the one-stage matching
    method performs poorly  
on the \emph{Kinect-sfm} benchmark, with RR decreasing by 25.0 pp. This proves the necessity 
of our prior-guided global dense matching and 
the superiority of two-stage matching.
More detailed ablation studies are provided in Appendix.

\section{Conclusion}

We propose a novel cross-source point cloud registration framework. With the density-robust feature extractor, we address the issue of inconsistent data distribution among cross-source point clouds. To tackle the challenging matching problem in registration, we employ a ``loose generation, strict selection'' strategy. Utilizing a two-stage matching approach, we obtain reliable correspondences, achieving robust registration using SVD. Future work will focus on developing a more comprehensive cross-source benchmark.

\section{Acknowledgements}
This work was funded by the National Key Research and Development Plan of China (No. 2018AAA0101000) and the National Natural Science Foundation of China under grant 62473052.

\bibliography{main}

\end{document}